\documentclass[conference]{IEEEtran}
\IEEEoverridecommandlockouts
\usepackage{amsmath,amssymb,amsfonts}
\usepackage{algorithm}
\usepackage[noend]{algpseudocode}
\usepackage{array}
\usepackage{svg}       
\usepackage[font=footnotesize,labelfont=rm,textfont=rm]{subcaption}
\usepackage{textcomp}
\usepackage{url}
\usepackage{verbatim}
\usepackage{graphicx}
\usepackage{cite}      
\usepackage{balance}    
\usepackage{xcolor}    
\usepackage{placeins} 
\usepackage{dblfloatfix}
\usepackage{booktabs}
\usepackage{multirow}
\def\BibTeX{{\rm B\kern-.05em{\sc i\kern-.025em b}\kern-.08em
    T\kern-.1667em\lower.7ex\hbox{E}\kern-.125emX}}

\algrenewcommand\algorithmicfor{\textbf{for}}
\algrenewcommand\algorithmicdo{\textbf{do}}
\algrenewcommand\algorithmicend{\textbf{end}}
\algrenewcommand\algorithmicwhile{\textbf{while}}
\algrenewcommand\algorithmicif{\textbf{if}}
\algrenewcommand\algorithmicthen{\textbf{then}}
\algrenewcommand\algorithmicelse{\textbf{else}}

\begin{document}
\title{Distributed LLM Inference via Communication-Efficient Speculative Decoding}

\author{
\IEEEauthorblockN{Joohyun~Shin, Byeonghyeon~Park, Keumseo~Ryum, and Joonhyuk~Kang}
\IEEEauthorblockA{School of Electrical Engineering, Korea Advanced Institute of Science and Technology (KAIST), Daejeon, South Korea\\
Email: \{shinjh1220, bhbahk, keumseo\}@kaist.ac.kr, jhkang@kaist.edu}
\thanks{\textit{J. Shin and B. Park contributed equally to this work.}}
}

\maketitle
\begin{abstract}
Distributed speculative decoding has emerged as a promising approach for device--server collaborative large language model (LLM) inference, but its decoding speedup is often limited by the vocabulary-scale communication overhead. In existing distributed speculative decoding methods, exact rejection sampling requires transmitting full token distributions over the LLM vocabulary between the device and the server, which becomes costly in bandwidth-constrained environments. To address this issue, we propose DISCO, a communication-efficient framework for distributed speculative decoding, that eliminates vocabulary-scale distribution transfer while preserving the exact target distribution. DISCO achieves this by introducing a server-side replica of the draft small language model (SLM), which enables the server to locally obtain the draft-side distribution required for rejection sampling. As a result, DISCO exchanges only drafted token indices and scalar probability values between the device and the server. Experimental results on multiple datasets show that DISCO consistently achieves higher decoding speedup than existing distributed speculative decoding methods, particularly under stringent bandwidth constraints.
\end{abstract}

\begin{IEEEkeywords}
LLM, speculative decoding, distributed inference, communication latency
\end{IEEEkeywords}

%
\IEEEpeerreviewmaketitle

\section{Introduction}

\IEEEPARstart{L}{arge} language models (LLMs) have achieved remarkable performance across a wide range of tasks, including reasoning, summarizing, and code generation \cite{gpt4techreport,llmsurvey}. Despite these advances, their autoregressive nature remains a key bottleneck for real-time applications. As each token is generated sequentially through a full forward pass of the model, the direct deployment of LLMs in bandwidth-constrained scenarios induces high latency. To address this issue without compromising generation quality, speculative decoding (SD) \cite{leviathan23a} has emerged as a promising solution. SD employs a small language model (SLM) to rapidly generate multiple draft tokens, which are then verified in parallel by the target LLM in a single forward pass. If a draft token is rejected, the target model resamples a corrected token from a modified distribution, thus preserving the exact distribution of the target model.

Recent advances in on-device AI accelerators have enabled SLMs to run directly on edge devices, thereby enabling collaborative deployment of language models in distributed environments \cite{edgeshard,edgeaisurvey}. In such a deployment, a device-side SLM can generate draft tokens while a server-side LLM performs verification, forming a distributed speculative decoding framework. This design allows the computational workload of LLM inference to be naturally partitioned between the device and the server, enabling efficient utilization of heterogeneous resources while maintaining the accuracy of the target model.

However, existing distributed SD methods  incur significant communication overhead.
In \cite{DSD}, the device transmits the full draft-token distributions over the entire vocabulary to the server for verification. Since modern LLMs typically use large vocabularies \cite{llama2,gemma}, transmitting these distributions incurs substantial communication overhead. To alleviate this issue, DSSD~\cite{dssd} shifts the resampling procedure to the device. During drafting, the device sends only the drafted token indices and their scalar probabilities instead of full draft-token distributions. Upon rejection, the server must still transmit the full target-token distribution to the device, which remains costly in bandwidth-constrained environments. 

To address these limitations, we propose DISCO, a distributed speculative decoding framework that preserves exact target sampling while removing vocabulary-dependent communication from the resampling procedure. The key idea is to make the draft-side distribution required for rejection sampling locally available at the server, thereby avoiding vocabulary-scale distribution transfer between the device and the server. To this end, DISCO uses a server-side replica of the draft SLM to reproduce the draft-side distribution upon rejection. As a result, the communication payload becomes independent of the vocabulary size and consists only of token indices and scalar probability values. The main contributions of this paper are summarized as follows:
\begin{itemize}
    \item We propose DISCO, a communication-efficient distributed speculative decoding framework that preserves exact target sampling while eliminating vocabulary-scale communication from rejection sampling. DISCO achieves this by localizing the draft-side distribution needed for resampling at the server through a replicated draft SLM.
    \item We derive an analytical formulation for the latency of distributed speculative decoding that characterizes the communication--computation trade-off, from which we identify the break-even bandwidth where DISCO becomes advantageous.
    
    \item 
    Through experiments under diverse bandwidth conditions, we demonstrate that DISCO consistently achieves higher speedup than existing distributed speculative decoding methods, with especially pronounced gains in low-bandwidth regimes.
\end{itemize}

\section{Background on Distributed SD}
We denote the device-side draft SLM and the target LLM by $M_q^d$ and $M_p$. We assume that they share a common vocabulary $\mathcal{V}$ of size $|\mathcal{V}|$.

\subsection{Speculative decoding}
\label{sec:spec_decoding}
Speculative decoding aims to reduce the latency of LLM inference by employing a smaller draft model to generate multiple tokens and verify them with the target LLM.  
\subsubsection{Draft Phase} 
The draft SLM generates $\gamma$ draft tokens $x_t, \ldots, x_{t+\gamma-1}$ autoregressively based on the current prefix, which consists of the input prompt and all previously generated tokens. We refer to $\gamma$ as the draft length, i.e., the number of candidate tokens generated in one speculative decoding round. 
At each step $i$, the next token
is sampled from $q_i=[q_{i,1},q_{i,2},\ldots,q_{i,|\mathcal{V}|}]^{T}$, where $q_i$ denotes the next-token distribution of the draft SLM.

\subsubsection{Verification Phase} 
Let $p_i = [p_{i,1}, p_{i,2}, \ldots, p_{i,|\mathcal{V}|}]^{T}$ denote the next-token distribution of the target LLM corresponding to the $i$-th token. For a drafted token $x_j$, let $q_j(x_j)$ and $p_j(x_j)$ denote its probabilities under $q_j$ and $p_j$, respectively. The token $x_j$ is accepted with probability $\min\!\left(1, {p_j(x_j)}/{q_j(x_j)}\right)$. Otherwise, a replacement token is resampled from $\mathrm{norm}(\max(0, p_j(x)-q_j(x)))$. If all $\gamma$ drafted tokens are accepted, the target LLM samples one additional token $x_{\gamma+1} \sim p_{\gamma+1}(x)$. This verification procedure ensures that the generated tokens follow the target LLM distribution.

\subsection{Distributed speculative decoding } 
\begin{figure}[t]
    \centering
    \includegraphics[width=0.9\linewidth]{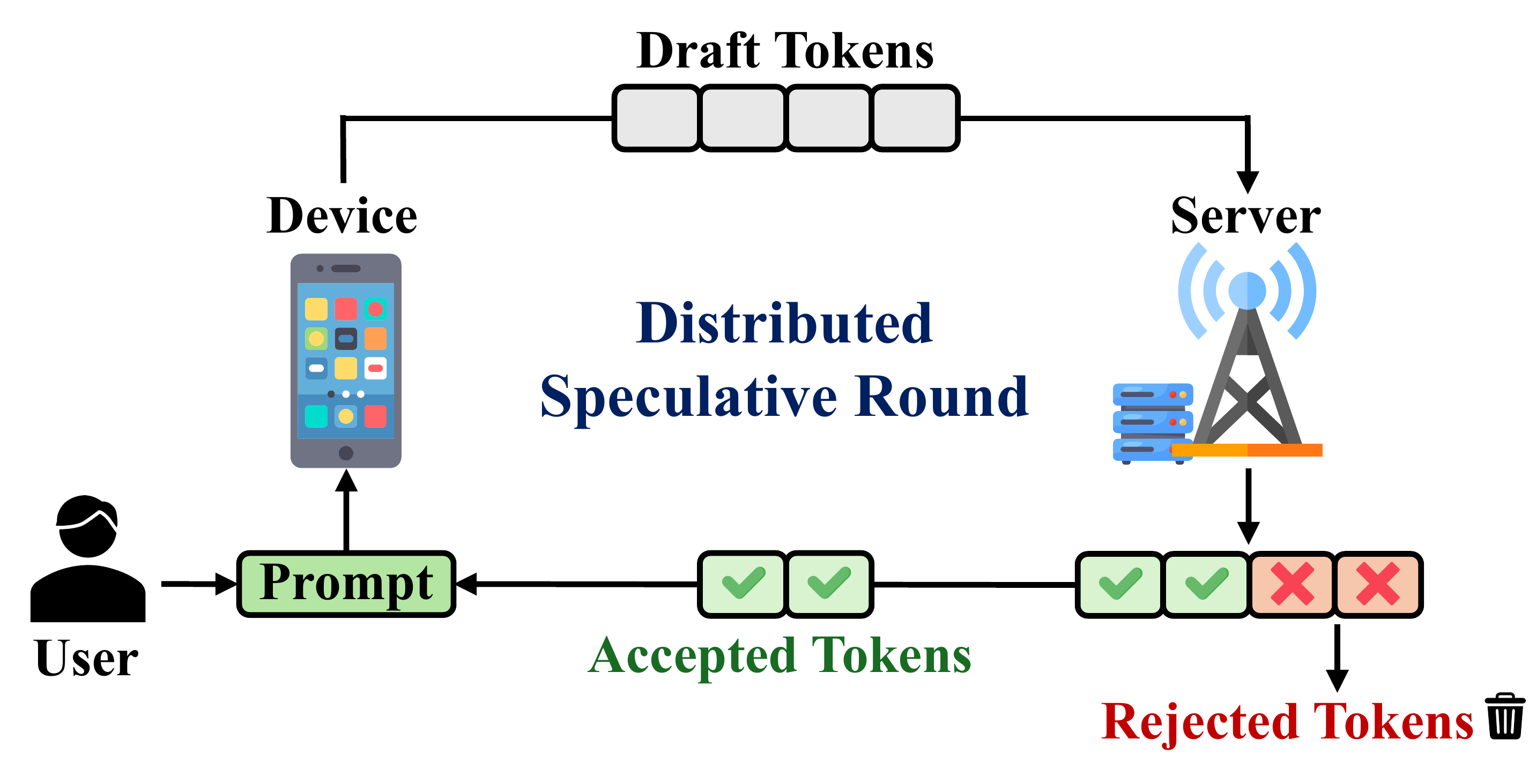}
    \caption{Distributed speculative decoding process. The device generates draft tokens using a SLM and sends them to the server. The server-side LLM verifies the received tokens and updates the prefix with the accepted tokens.}
    \label{fig:Distributed Setting}
\end{figure}

Fig.~\ref{fig:Distributed Setting} illustrates the distributed SD framework considered in this work. In this setting, the drafting phase is offloaded to the device equipped with a SLM, while the verification phase is performed at the server hosting the target LLM. The device SLM generates $\gamma$ draft tokens and transmits them to the server, where the target LLM verifies the tokens received according to the SD procedure described in Section~\ref{sec:spec_decoding}. Accepted tokens are appended to the prefix, while rejected tokens are discarded and replaced through a resampling step. Prior methods on distributed speculative decoding mainly differ in where the resampling step is performed and what information must be exchanged between the device and the server.


\subsubsection{DSD} 
We denote the preliminary work \cite{DSD} on SD in distributed environments as DSD. In the draft phase of DSD, the draft SLM on the device generates draft tokens autoregressively. The target LLM on the server performs both verification and resampling. This requires the device to send the drafted token indices together with their full draft-token distributions. While this design preserves the original speculative decoding procedure, it incurs substantial uplink communication overhead due to the transmission of full draft-token  distributions. 

\subsubsection{DSSD} 
 To reduce this cost, DSSD~\cite{dssd} shifts the resampling step to the device. In this approach, the device sends only the token indices and their scalar probabilities $q_i(x_i)$ after the drafting phase, while the server transmits the full target-token distribution $p_j$ only when a rejection occurs at position $j$. As a result, DSSD reduces uplink communication during drafting, but still requires transmitting the full target-token distribution upon rejection.
\begin{algorithm}[hbt!]
\caption{Pseudocode of DISCO}
\label{alg:disco}
\begin{algorithmic}[1]

\State \textbf{Draft Phase (Device)}
\State \textbf{Input:} $M_q^{d}$, \textit{prefix}
\For{$i = 1$ \textbf{to} $\gamma$}
    \State $q_i(x) \gets M_q^{d}(\textit{prefix} + [x_1,\cdots,x_{i-1}])$
    \State $x_i \sim q_i(x)$
\EndFor

\State \textbf{Uplink Transmission}
\State Send $[x_1, \dots, x_\gamma]$ and $[q_1(x_1), \dots, q_\gamma(x_\gamma)]$ to Server

\State \textbf{Verification Phase (Server)}
\State \textbf{Input:} $M_p$, $M_q^s$, $\textit{prefix}$, $[x_1, \dots, x_\gamma]$, $[q_1(x_1), \dots, q_\gamma(x_\gamma)]$

\State \textbf{Initialize:} $j = 1$
\State $p_1(x), \ldots, p_{\gamma+1}(x) \gets$
\Statex \hspace{\algorithmicindent} $M_p(\textit{prefix}), \ldots, M_p(\textit{prefix} + [x_1, \ldots, x_\gamma])$
\While{$j \le \gamma$}
    \State $r_j \sim U[0, 1]$ 
    \If{$r_j \leq \min\left\{1, {p_j(x_j)}/{q_j(x_j)}\right\}$ } \Comment{Accept}
        \State $j \gets j + 1$
    \Else \Comment{Reject}
        \State $q'_j(x) \gets M^{s}_{q}(\textit{prefix} + [x_1, \dots, x_{j-1}])$
        \State $x'_j \sim \text{norm}(\max(0, p_j(x) - q'_j(x)))$ \Comment{Resample}
        \State $x_j \gets x'_j$
        \State \textbf{Break}
    \EndIf
\EndWhile 
\If{$j = \gamma + 1$ }
    \State $x_{\gamma+1} \sim p_{\gamma+1}(x)$
\EndIf
\State $\textit{prefix} \gets \textit{prefix} + [x_1, \dots, x_j]$ \Comment{Reset $M_q^s$ prefix}
\State \textbf{Downlink Transmission}
\State Send token $x_j$ and position $j$ to Device
\State $\textit{prefix} \gets \textit{prefix} + [x_1, \dots, x_j]$ \Comment{Reset $M_q^d$ prefix}
\end{algorithmic}
\end{algorithm}

\section{DISCO}
\begin{figure*}[t]
    \centering
    \includegraphics[width=\linewidth]{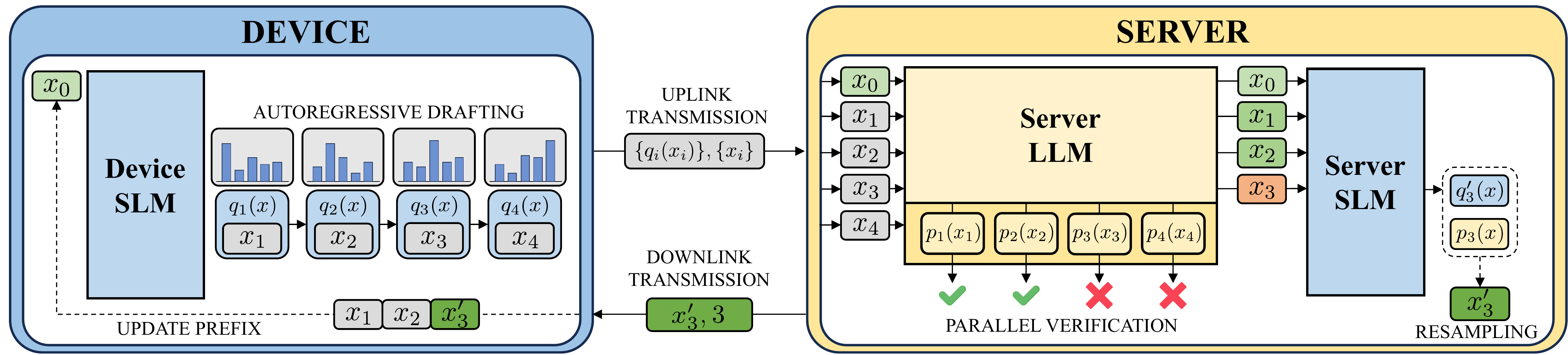}
    \caption{Framework of DISCO. The device generates draft tokens and sends only token indices and scalar probabilities to the server. The server performs parallel verification using the LLM, and upon rejection, conducts server-side resampling with the server SLM. The accepted prefix and the resampled token are then fed back to the device.}
    \label{fig:schemes}
\end{figure*}
DISCO addresses the vocabulary-scale communication overhead of prior distributed speculative decoding methods by performing resampling locally at the server, as illustrated in Fig.~\ref{fig:schemes}. To this end, DISCO introduces an additional server-side draft SLM $M_q^s$, which is identical to the device-side draft SLM $M_q^d$. We assume that $M_q^d$ and $M_q^s$ share the same tokenizer, model parameters, and decoding pipeline, so that they produce matched next-token distributions for the same prefix. Upon rejection, the server uses $M_q^s$ to reproduce the draft-side distribution at the rejected position through a single forward pass, instead of receiving a full token distribution from the device. The server then performs resampling locally using the reproduced draft-side distribution and the target-token distribution from $M_p$. If all draft tokens are accepted, the server-side LLM samples an additional token and sends it back to the device. The overall process of DISCO is summarized in Algorithm~\ref{alg:disco}.

\section{Communication Cost and Latency Analysis}

We first quantify the communication payload of DSD, DSSD, and DISCO,
and then characterize how the payload reduction translates into
per-round latency. The three methods differ primarily in where
resampling is performed and, consequently, whether a vocabulary-scale
token distribution must be transmitted over the uplink or downlink.

Let $b_{\mathrm{tok}}$, $b_{\mathrm{prob}}$, and $b_{\mathrm{pos}}$
denote the bitwidths of a token index, a probability value, and a
position index, respectively. We denote by $A_\gamma$ the probability that all $\gamma$ draft tokens are accepted in a speculative decoding round.

Table~\ref{tab:comm_payload} summarizes the communication payload of
one speculative decoding round. In DSD, the device transmits the full
draft-token distributions over the uplink. DSSD avoids this
vocabulary-scale uplink transmission by sending only the drafted token
indices and their scalar probabilities, but requires the server to
transmit a full target-token distribution over the downlink when a
rejection occurs. In contrast, DISCO avoids vocabulary-scale
transmission in both directions. In our implementation, token and position indices are represented using 32 bits, while probability values use 16 bits. For OPT
($|\mathcal{V}|=50{,}272$) with $\gamma=4$, DSD requires approximately
402.2 kB per round on the uplink, whereas DISCO requires only 24 B.
Upon rejection, DSSD requires approximately 100.5 kB on the downlink,
whereas DISCO requires only 8 B.

\begin{table}[t]
\caption{Communication payload per speculative decoding round.}
\label{tab:comm_payload}
\centering
\resizebox{\columnwidth}{!}{
\begin{tabular}{lcc}
\toprule
Method & Uplink payload & Expected downlink payload \\
\midrule
DSD
& $\gamma\!\left(|\mathcal{V}|b_{\mathrm{prob}}+b_{\mathrm{tok}}\right)$
& $b_{\mathrm{tok}}+b_{\mathrm{pos}}$ \\

DSSD
& $\gamma\!\left(b_{\mathrm{tok}}+b_{\mathrm{prob}}\right)$
& $b_{\mathrm{pos}}
+A_\gamma b_{\mathrm{tok}}
+(1-A_\gamma)|\mathcal{V}|b_{\mathrm{prob}}$ \\

DISCO
& $\gamma\!\left(b_{\mathrm{tok}}+b_{\mathrm{prob}}\right)$
& $b_{\mathrm{tok}}+b_{\mathrm{pos}}$ \\
\bottomrule
\end{tabular}}
\end{table}

We next translate these communication payloads into latency.
Let $C_{\mathrm{UL}}$ and $C_{\mathrm{DL}}$ denote the effective
uplink and downlink data rates in bits/s, respectively. The
communication latency of one speculative decoding round is modeled as
\begin{equation}
    T_{\mathrm{comm}}
    =
    RTT
    +
    \frac{B_{\mathrm{up}}}{C_{\mathrm{UL}}}
    +
    \frac{B_{\mathrm{down}}}{C_{\mathrm{DL}}},
    \label{comm_time}
\end{equation}
where $B_{\mathrm{up}}$ and $B_{\mathrm{down}}$ denote the uplink and
downlink payloads in bits. Hence, each payload-dependent transmission
term has units of seconds. The $RTT$ term represents the
bandwidth-independent network round-trip delay, whereas
$B/C$ represents the payload-dependent transmission delay.

For the symmetric network setting considered in our main experiments,
we assume
\begin{equation}
    C_{\mathrm{UL}} = C_{\mathrm{DL}} = BW.
\end{equation}

For all methods, the base latency of a speculative decoding round consists
of the device-side drafting time $T_d(\gamma)$, the server-side
verification time $T_v(\gamma)$, and the communication latency:
\begin{equation}
    T_{\mathrm{round}}
    =
    T_d(\gamma)
    +
    T_v(\gamma)
    +
    T_{\mathrm{comm}}.
    \label{round_time}
\end{equation}

We further denote by $T_s$ the additional server-side SLM computation
time incurred by DISCO when rejection and resampling occur. Using the
payloads in Table~\ref{tab:comm_payload}, the expected round latency of
DSD is
\begin{equation}
\begin{aligned}
\mathbb{E}\!\left[T_{\mathrm{round}}^{\mathrm{DSD}}\right]
=&\;
T_d(\gamma)
+
T_v(\gamma)
+
RTT \\
&+
\frac{
\gamma\!\left(
|\mathcal{V}|b_{\mathrm{prob}}+b_{\mathrm{tok}}
\right)
+
b_{\mathrm{tok}}
+
b_{\mathrm{pos}}
}{BW}.
\end{aligned}
\label{DSD_round}
\end{equation}

\begin{figure*}[!t]
    \centering
    \begin{subfigure}[b]{0.32\textwidth}
        \centering
        \captionsetup{margin={20pt, 0pt}}
        \includegraphics[width=\textwidth]{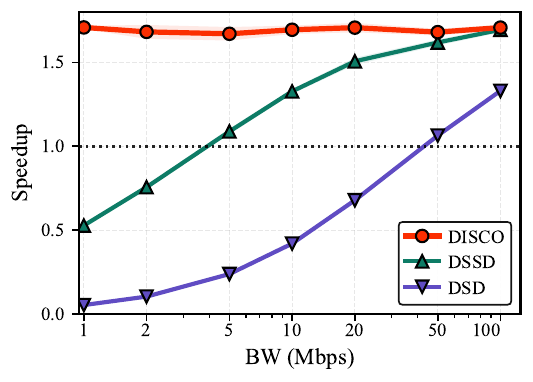}
        \caption{GSM8K}
        \label{fig:bw_gsm8k}
    \end{subfigure}
    \hfill
    \begin{subfigure}[b]{0.32\textwidth}
        \centering
        \captionsetup{margin={20pt, 0pt}}
        \includegraphics[width=\textwidth]{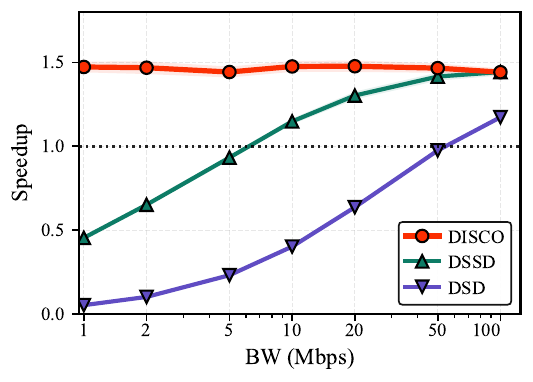}
        \caption{CNN/DailyMail}
        \label{fig:bw_cnn}
    \end{subfigure}
    \hfill
    \begin{subfigure}[b]{0.32\textwidth}
        \centering
        \captionsetup{margin={20pt, 0pt}}
        \includegraphics[width=\textwidth]{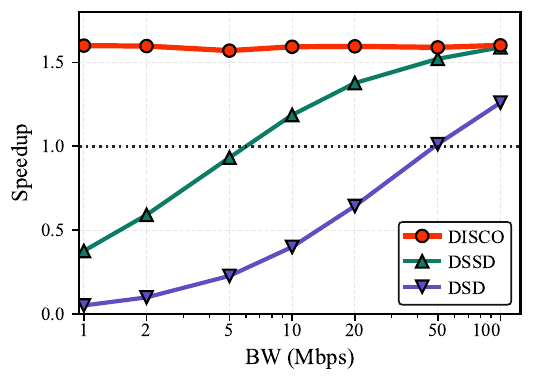}
        \caption{HumanEval}
        \label{fig:bw_humaneval}
    \end{subfigure}
    \caption{Speedup of DISCO, DSSD, and DSD under varying bandwidth across different datasets with $RTT$ fixed to 20 ms. DISCO consistently outperforms the baselines across all datasets and bandwidth settings, with especially clear benefits in the low-bandwidth regime.}
    \label{fig:bw_sweep}
\end{figure*}

For DSSD, the device sends $\gamma$ token indices and their scalar
probabilities over the uplink. On the downlink, the server sends an
additional token when all draft tokens are accepted, whereas it sends
the full target-token distribution when a rejection occurs. Therefore,
\begin{equation}
\begin{aligned}
\mathbb{E}\!\left[T_{\mathrm{round}}^{\mathrm{DSSD}}\right]
=&\;
T_d(\gamma)
+
T_v(\gamma)
+
RTT \\
&+
\frac{
\gamma\!\left(
b_{\mathrm{tok}}+b_{\mathrm{prob}}
\right)
+
b_{\mathrm{pos}}
+
A_\gamma b_{\mathrm{tok}}
}{BW} \\
&+
\frac{
(1-A_\gamma)
|\mathcal{V}|b_{\mathrm{prob}}
}{BW}.
\end{aligned}
\label{DSSD_round}
\end{equation}

For DISCO, both uplink and downlink payloads are independent of the
vocabulary size. Its expected round latency is
\begin{equation}
\begin{aligned}
\mathbb{E}\!\left[T_{\mathrm{round}}^{\mathrm{DISCO}}\right]
=&\;
T_d(\gamma)
+
T_v(\gamma)
+
RTT \\
&+
\frac{
\gamma\!\left(
b_{\mathrm{tok}}+b_{\mathrm{prob}}
\right)
+
b_{\mathrm{tok}}
+
b_{\mathrm{pos}}
}{BW} \\
&+
(1-A_\gamma)T_s.
\end{aligned}
\label{DISCO_round}
\end{equation}

The last term in (\ref{DISCO_round}) accounts for the additional
server-side SLM computation required for resampling. Since this
computation is invoked only when at least one of the $\gamma$ draft
tokens is rejected, it occurs with probability $1-A_\gamma$.

Equations~(\ref{DSD_round})--(\ref{DISCO_round}) directly correspond
to the message exchanges summarized in Table~\ref{tab:comm_payload}.
Compared with DSD, DISCO eliminates vocabulary-scale communication
from the uplink, while compared with DSSD, it eliminates the
vocabulary-scale target distribution from the downlink upon rejection.

Since DSSD and DISCO have the same uplink payload, their latency
difference is determined by the downlink transmission upon rejection
and the additional server-side SLM computation. Assuming
$b_{\mathrm{tok}} \ll |\mathcal{V}|b_{\mathrm{prob}}$, we obtain
\begin{equation}
\mathbb{E}[T_{\mathrm{round}}^{\mathrm{DSSD}}]
-
\mathbb{E}[T_{\mathrm{round}}^{\mathrm{DISCO}}]
\simeq
(1-A_\gamma)
\left(
\frac{|\mathcal{V}|b_{\mathrm{prob}}}{BW}
-
T_s
\right).
\label{latency_difference}
\end{equation}

Since $1-A_\gamma>0$, the approximate break-even bandwidth is
\begin{equation}
    BW^*
    \simeq
    \frac{|\mathcal{V}|b_{\mathrm{prob}}}{T_s},
    \label{breakeven}
\end{equation}
such that DISCO achieves a lower expected round latency than DSSD when
$BW<BW^*$. Moreover, $BW^*$ increases approximately linearly with
$|\mathcal{V}|$, indicating that DISCO remains advantageous over a
wider range of bandwidth conditions as the vocabulary size grows.

\section{Experimental Results}
\subsection{Experimental Setup}
We use OPT-125m \cite{opt} as the draft model and OPT-30B as the target model, both in fp16 precision. 
To further examine whether the observed trend generalizes across model families, we additionally evaluate Qwen2.5-0.5B \cite{qwen} as the draft model and Qwen2.5-32B as the target model in Table~\ref{tab:speedup_bw_gamma} . 
All experiments are conducted on two NVIDIA A100 40GB GPUs. The draft model runs on a single GPU, whereas the target model is distributed across both GPUs due to its memory requirement. We evaluate our method on three datasets, GSM8K \cite{gsm8k}, CNN/DailyMail \cite{cnn}, and HumanEval \cite{humaneval}.

As the main evaluation metric, we measure speedup, defined as the average throughput relative to vanilla autoregressive decoding. For each dataset, we evaluate 100 samples and report the average speedup across samples. 
We simulate network conditions by varying bandwidth $BW$ from 1 to 100 Mbps and draft length $\gamma$ from 2 to 8, while fixing $RTT$ to 20 ms. To reduce randomness, all results are averaged over three random seeds. 
We compare DISCO with DSD \cite{DSD} and DSSD \cite{dssd}, two representative baselines for distributed speculative decoding that also preserve the exact target distribution. 

\begin{table*}[t]
\vspace{0.04in}
\centering
\small
\caption{Speedup on GSM8K with varying draft length under different bandwidth conditions over 3 seeds. BW is measured in Mbps. For each model, bandwidth, and $\gamma$, the best speedup among DSD, DSSD, and DISCO is in bold.}
\label{tab:speedup_bw_gamma}
\begin{tabular}{llcccccccccccc}
\toprule
& & \multicolumn{3}{c}{$\gamma=2$}
& \multicolumn{3}{c}{$\gamma=4$}
& \multicolumn{3}{c}{$\gamma=6$}
& \multicolumn{3}{c}{$\gamma=8$} \\
\cmidrule(lr){3-5}
\cmidrule(lr){6-8}
\cmidrule(lr){9-11}
\cmidrule(lr){12-14}
Model & BW
& DSD & DSSD & DISCO
& DSD & DSSD & DISCO
& DSD & DSSD & DISCO
& DSD & DSSD & DISCO \\
\midrule

\multirow{4}{*}{OPT-125m/30B}
& 1
& 0.08 & 0.57 & \textbf{1.41}
& 0.05 & 0.53 & \textbf{1.67}
& 0.04 & 0.56 & \textbf{1.81}
& 0.04 & 0.56 & \textbf{1.85} \\

& 5
& 0.31 & 1.01 & \textbf{1.40}
& 0.24 & 1.09 & \textbf{1.67}
& 0.20 & 1.17 & \textbf{1.81}
& 0.18 & 1.19 & \textbf{1.86} \\

& 10
& 0.51 & 1.18 & \textbf{1.41}
& 0.42 & 1.33 & \textbf{1.69}
& 0.37 & 1.43 & \textbf{1.84}
& 0.32 & 1.46 & \textbf{1.89} \\

& 50
& 1.06 & 1.38 & \textbf{1.42}
& 1.06 & 1.62 & \textbf{1.68}
& 1.02 & 1.74 & \textbf{1.82}
& 0.94 & 1.74 & \textbf{1.84} \\
\midrule

\multirow{4}{*}{Qwen2.5-0.5B/32B}
& 1
& 0.04 & 0.44 & \textbf{1.34}
& 0.03 & 0.40 & \textbf{1.55}
& 0.02 & 0.39 & \textbf{1.62}
& 0.02 & 0.39 & \textbf{1.63} \\

& 5
& 0.18 & 0.93 & \textbf{1.34}
& 0.13 & 0.96 & \textbf{1.56}
& 0.12 & 1.00 & \textbf{1.63}
& 0.11 & 0.98 & \textbf{1.63} \\

& 10
& 0.30 & 1.08 & \textbf{1.33}
& 0.26 & 1.21 & \textbf{1.56}
& 0.21 & 1.21 & \textbf{1.62}
& 0.19 & 1.20 & \textbf{1.62} \\

& 50
& 0.79 & 1.29 & \textbf{1.34}
& 0.75 & 1.48 & \textbf{1.56}
& 0.72 & 1.55 & \textbf{1.63}
& 0.66 & 1.55 & \textbf{1.63} \\
\bottomrule
\end{tabular}
\end{table*}
\subsection{Performance Under Varying Network Conditions}
We first analyze the impact of network conditions on decoding performance across different datasets. Fig.~\ref{fig:bw_sweep} shows the speedup achieved by DISCO, DSSD, and DSD under varying bandwidth conditions for GSM8K, CNN/DailyMail, and HumanEval, with the draft length fixed at $\gamma=4$. Across all datasets, DISCO consistently maintains stable performance and achieves approximately 1.5$\times$ speedup even under extreme bandwidth restrictions. In contrast, both DSD and DSSD are highly sensitive to bandwidth limitations and exhibit significant performance degradation in low-bandwidth regimes. This behavior stems from the high communication overhead for verification, since DSD and DSSD require transmitting token distributions between the device and the server. By transmitting only a small number of candidate tokens and scalar values, DISCO substantially reduces the communication cost, enabling robust performance even in bandwidth-constrained environments.

Table~\ref{tab:speedup_bw_gamma} reports the speedup on GSM8K under different draft lengths $\gamma$, bandwidths, and model families. DISCO achieves the best speedup in all evaluated settings, showing that the proposed communication-efficient design is robust to bandwidth variation and the choice of the model. The performance gap becomes more pronounced under bandwidth-constrained regimes ($BW=1$ Mbps, $BW=5$ Mbps), as DISCO does not require vocabulary-scale distribution transfer during drafting or upon rejection, maintaining more than $1.3\times$ speedup for both models.

The results also show different sensitivity to the draft length $\gamma$. For DSD, speedup consistently decreases as $\gamma$ increases, since a larger $\gamma$ requires transmitting more full draft-token distributions. In contrast, DISCO maintains high speedup across different values of $\gamma$, especially for $\gamma=4$--$8$. This indicates that eliminating vocabulary-scale communication makes DISCO less sensitive to the draft length and enables stable acceleration even under limited bandwidth.

\subsection{Communication--Computation Trade-off Analysis}
\begin{figure}[t]
    \centering
    \includegraphics[width=\linewidth]{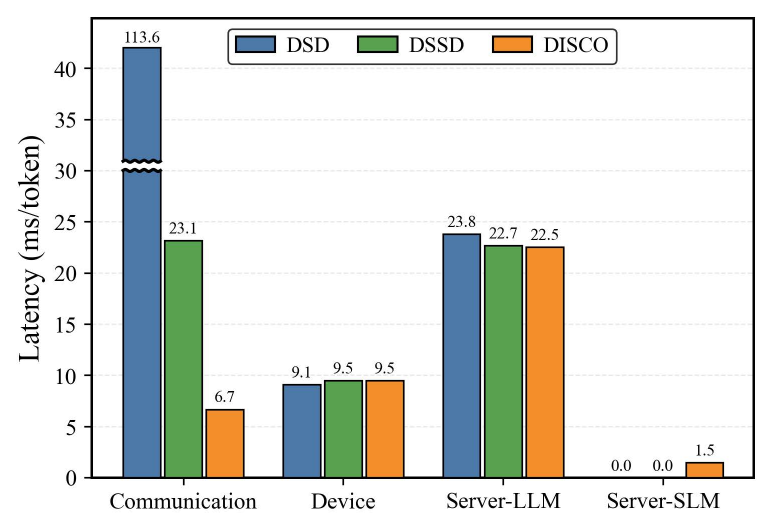}
    \caption{Latency breakdown per token under $BW=10$ Mbps and $RTT=20$ ms on GSM8K. DISCO significantly reduces communication latency compared to DSD and DSSD, while computation latency remains similar across methods.}
    \label{fig:latency_breakdown}
\end{figure}

To identify the factors that contribute to these performance differences, we decompose the latency of each method. Fig.~\ref{fig:latency_breakdown} shows the per-token latency components, including communication, device computation, and server-side computation. The major difference between methods arises from communication latency, which is significantly larger for DSD and DSSD. In contrast, DISCO greatly reduces this overhead by transmitting only a small number of tokens and probability values. 

Notably, the computation latency remains similar across methods. The server-side SLM computation introduces negligible overhead since it performs a single forward pass only when resampling is required upon rejection. Specifically, our profiling results show that the target LLM consumes approximately 143.36 GFLOPs/token, whereas the additional server-side SLM of DISCO requires only 0.36 GFLOPs/token. This indicates that the computational burden added by DISCO is marginal. Minor differences in device- and server-side LLM latency arise from the location of the resampling step for each method. These results indicate that the primary bottleneck in distributed speculative decoding is communication rather than computation, and the performance gain of DISCO mainly comes from reducing the communication cost. 

Using the latency breakdown in Fig.~\ref{fig:latency_breakdown} for the GSM8K dataset, we can approximate the parameters in (\ref{latency_difference}) and estimate the empirical break-even bandwidth. The communication term approximately scales with $1/BW$, while the device- and server-side computation times remain nearly constant. As shown in Fig.~\ref{fig:latency_breakdown}, the bandwidth-independent computation time differs slightly between DSSD and DISCO due to the additional server-side resampling time in DISCO, yielding a difference of approximately 1.5 ms/token. At $BW=10$ Mbps, the difference in communication latency between DSSD and DISCO is about 16.4 ms/token. Based on (\ref{latency_difference}), the break-even bandwidth can be approximated as
\begin{equation}
    BW^*\approx\frac{16.4}{1.5}\times 10 \approx 109\text{ Mbps}.
\end{equation}
This prediction is consistent with the experimental results in Fig.~\ref{fig:bw_gsm8k}, where the speedup gap between the two methods becomes small in the high-bandwidth regime and the two curves exhibit similar speedups around 100 Mbps.

We note that this break-even bandwidth scales linearly with the LLM vocabulary size, as indicated by (\ref{breakeven}). Scaling law analyses have shown that compute-optimal vocabulary size increases with model scale, and that most existing LLMs employ vocabularies smaller than the optimal value~\cite{vocab}. As the communication payload of DISCO remains vocabulary-independent, the bandwidth regime where DISCO outperforms DSSD is expected to widen further as LLMs continue to adopt larger vocabularies.

\subsection{Heterogeneous Device–Server Deployment}
To emulate a realistic distributed environment with a resource-constrained device GPU and a relatively large server GPU, we consider heterogeneous GPU configurations in which the device uses either an RTX 3090 24GB or an RTX 4090 24GB, while the server is equipped with two NVIDIA A100 40GB GPUs. The GPUs are connected over a local area network (LAN) where the network conditions are measured as $RTT=0.5$ ms and $BW=84.3$ Mbps.

\begin{table}[t]
    \caption{Average decoding speedup across datasets under different device--server GPU configurations.}    
    \label{tab:speedup_real}
    \centering
    \setlength{\tabcolsep}{5pt}
    \renewcommand{\arraystretch}{1.12}
    \begin{tabular}{llccc}
    \toprule
    \multirow{2}{*}{\textbf{Config.}} & \multirow{2}{*}{\textbf{Method}} & \multicolumn{3}{c}{\textbf{Dataset}} \\
    \cmidrule(lr){3-5}
    & & \textbf{GSM8K} & \textbf{CNN/DM} & \textbf{HumanEval} \\
    \midrule
    \multirow{3}{*}{RTX3090--A100}
      & DISCO & \textbf{1.75} & \textbf{1.65} & \textbf{1.64} \\
      & DSSD  & 1.57          & 1.52          & 1.44          \\
      & DSD   & 0.92          & 0.92          & 0.86          \\
    \midrule
    \multirow{3}{*}{RTX4090--A100}
      & DISCO & \textbf{1.92} & \textbf{1.78} & \textbf{1.78} \\
      & DSSD  & 1.72          & 1.63          & 1.57          \\
      & DSD   & 0.97          & 0.96          & 0.91          \\
    \bottomrule
    \end{tabular}
\end{table}
 
Table~\ref{tab:speedup_real} reports the average decoding speedup across datasets under the two device--server configurations. DISCO achieves the highest speedup in all settings and reaches up to $1.92\times$ under the RTX 4090--NVIDIA A100 configuration. DSSD also improves over autoregressive decoding, but its gain remains smaller than that of DISCO. In contrast, DSD fails to outperform autoregressive decoding, empirically demonstrating that vocabulary-scale communication is a major bottleneck in distributed speculative decoding.

Compared with the simulation results in Fig.~\ref{fig:bw_sweep}, the measured speedups are slightly higher. This trend aligns with the relatively high bandwidth and low $RTT$ of the LAN environment. Overall, these results confirm that DISCO remains effective in practical heterogeneous device--server settings.

\section{Conclusion}
In this work, we proposed DISCO, a distributed speculative decoding framework for communication-efficient LLM inference. DISCO preserves the exact target distribution while eliminating vocabulary-scale distribution transfer through an additional server-side SLM for resampling. As only token indices and scalar probability values are transmitted, DISCO significantly reduces communication latency and data transfer overhead. Experiments in both simulation-based environments with varying bandwidth and a realistic distributed environment with heterogeneous GPUs demonstrate that DISCO achieves higher speedup than autoregressive generation and existing distributed speculative decoding methods. By modifying only the locality of the draft-side distribution required for resampling, rather than the underlying target model or exact acceptance rule, DISCO is naturally extensible to other model families and exact speculative decoding variants. While our current evaluation focuses on single-request settings, we leave the extension of DISCO to multi-user scenarios with concurrent requests for future work.

\ifCLASSOPTIONcaptionsoff
  \newpage
\fi

\bibliographystyle{IEEEtran}
\bibliography{refs}

\end{document}